%% file: main.tex
\documentclass[11pt,a4paper]{article}
\usepackage{naacl2021}
\usepackage{times}
\usepackage{latexsym}

\usepackage{microtype}
\input{settings.tex}

\input{math_commands.tex}

\newcommand\our{\textsc{InfoXLM}}

\newcommand\xlco{\textsc{XlCo}}

\newcommand{\tblidx}[1]{{\scriptsize \texttt{[#1]}}}

\title{\our{}: An Information-Theoretic Framework for \\ Cross-Lingual Language Model Pre-Training}

\author{Zewen Chi$^{\dag\ddag}$\thanks{\ \  Contribution during internship at Microsoft Research. Contact person: Li Dong and Furu Wei.
},~~Li Dong$^\ddag$,~~Furu Wei$^\ddag$,~~Nan Yang$^{\ddag}$,~~Saksham Singhal$^{\ddag}$,~~Wenhui Wang$^{\ddag}$\\
\textbf{Xia Song}$^{\ddag}$\textbf{,}~~\textbf{Xian-Ling Mao}$^\dag$\textbf{,}~~\textbf{Heyan Huang}$^\dag$\textbf{,}~~\textbf{Ming Zhou}$^\ddag$\\
$^\dag$Beijing Institute of Technology \\
$^\ddag$Microsoft Corporation \\
\texttt{\{czw,maoxl,hhy63\}@bit.edu.cn}
\\\texttt{\{lidong1,fuwei,nanya,saksingh,wenwan,xiaso\}@microsoft.com} \\}

\date{}

\begin{document}
\maketitle
\begin{abstract}
In this work, we present an information-theoretic framework that formulates cross-lingual language model pre-training as maximizing mutual information between multilingual-multi-granularity texts. The unified view helps us to better understand the existing methods for learning cross-lingual representations. More importantly, inspired by the framework, we propose a new pre-training task based on contrastive learning. Specifically, we regard a bilingual sentence pair as two views of the same meaning and encourage their encoded representations to be more similar than the negative examples. By leveraging both monolingual and parallel corpora, we jointly train the pretext tasks to improve the cross-lingual transferability of pre-trained models. Experimental results on several benchmarks show that our approach achieves considerably better performance. The code and pre-trained models are available at \url{https://aka.ms/infoxlm}.
\end{abstract}

\section{Introduction}
\label{sec:intro}

Learning cross-lingual language representations plays an important role in overcoming the language barrier of NLP models.
The recent success of cross-lingual language model pre-training~\cite{bert,xlm,xlmr,xnlg,mbart} significantly improves the cross-lingual transferability in various downstream tasks, such as cross-lingual classification, and question answering.

State-of-the-art cross-lingual pre-trained models are typically built upon multilingual masked language modeling (MMLM;~\citealt{bert,xlmr}), and translation language modeling (TLM;~\citealt{xlm}).
The goal of both pretext tasks is to predict masked tokens given input context. The difference is that MMLM uses monolingual text as input, while TLM feeds bilingual parallel sentences into the model.
Even without explicit encouragement of learning universal representations across languages, the derived models have shown promising abilities of cross-lingual transfer.

In this work, we formulate cross-lingual pre-training from a unified information-theoretic perspective.
Following the mutual information maximization principle~\cite{infomax,infoword}, we show that the existing pretext tasks can be viewed as maximizing the lower bounds of mutual information between various multilingual-multi-granularity views.

Specifically, MMLM maximizes mutual information between the masked tokens and the context in the same language while the anchor points across languages encourages the correlation between cross-lingual contexts.
Moreover, we present that TLM can maximize mutual information between the masked tokens and the parallel context, which implicitly aligns encoded representations of different languages.
The unified information-theoretic framework also inspires us to propose a new cross-lingual pre-training task, named as cross-lingual contrast (\xlco{}).
The model learns to distinguish the translation of an input sentence from a set of negative examples. 
In comparison to TLM that maximizes token-sequence mutual information, \xlco{} maximizes sequence-level mutual information between translation pairs which are regarded as cross-lingual views of the same meaning.
We employ the momentum contrast~\cite{moco} to realize \xlco{}.
We also propose the mixup contrast and conduct the contrast on the universal layer to further facilitate the cross-lingual transferability.

Under the presented framework, we develop a cross-lingual pre-trained model (\our{}) to leverage both monolingual and parallel corpora.
We jointly train \our{} with MMLM, TLM and \xlco{}.
We conduct extensive experiments on several cross-lingual understanding tasks, including cross-lingual natural language inference~\cite{xnli}, cross-lingual question answering~\cite{mlqa}, and cross-lingual sentence retrieval~\cite{tatoeba}.
Experimental results show that \our{} outperforms strong baselines on all the benchmarks.
Moreover, the analysis indicates that \our{} achieves better cross-lingual transferability.

\section{Related Work}

\subsection{Cross-Lingual LM Pre-Training}

Multilingual BERT (mBERT;~\citealt{bert}) is pre-trained with the multilingual masked language modeling (MMLM) task on the monolingual text.
mBERT produces cross-lingual representations and performs cross-lingual tasks surprisingly well~\cite{wu2019beto}.
XLM~\cite{xlm} extends mBERT with the translation language modeling (TLM) task so that the model can learn cross-lingual representations from parallel corpora.
Unicoder~\cite{unicoder} tries several pre-training tasks to utilize parallel corpora.
ALM~\cite{alm} extends TLM to code-switched sequences obtained from translation pairs.
XLM-R~\cite{xlmr} scales up MMLM pre-training with larger corpus and longer training.
LaBSE~\cite{labse} learns cross-lingual sentence embeddings by an additive translation ranking loss.

In addition to learning cross-lingual encoders, several pre-trained models focus on generation.
MASS~\cite{mass} and mBART~\cite{mbart} pretrain sequence-to-sequence models to improve machine translation.
XNLG~\cite{xnlg} focuses on the cross-lingual transfer of language generation, such as cross-lingual question generation, and abstractive summarization.

\subsection{Mutual Information Maximization}

Various methods have successfully learned visual or language representations by maximizing mutual information between different views of input.
It is difficult to directly maximize mutual information.
In practice, the methods resort to a tractable lower bound as the estimator, such as InfoNCE~\cite{infonce}, and the variational form of the KL divergence~\cite{nwj}.
The estimators are also known as contrastive learning~\cite{Arora2019ATA} that measures the representation similarities between the sampled positive and negative pairs.
In addition to the estimators, various view pairs are employed in these methods.
The view pair can be the local and global features of an image~\cite{infomax,bachman2019learning}, the random data augmentations of the same image~\cite{cmc,moco,simclr}, or different parts of a sequence~\cite{infonce,cpc2,infoword}.
\citet{infoword} show that learning word embeddings or contextual embeddings can also be unified under the framework of mutual information maximization.

\section{Information-Theoretic Framework for Cross-Lingual Pre-Training}

In representation learning, the learned representations are expected to preserve the information of the original input data.
However, it is intractable to directly model the mutual information between the input data and the representations. Alternatively, we can maximize the mutual information between the representations from different views of the input data, e.g., different parts of a sentence, a translation pair of the same meaning.

In this section, we start from a unified information-theoretic perspective, and formulate cross-lingual pre-training with the mutual information maximization principle. Then, under the information-theoretic framework, we propose a new cross-lingual pre-training task, named as cross-lingual contrast (\xlco{}). Finally, we present the pre-training procedure of our \our{}.

\subsection{Multilingual Masked Language Modeling}
\label{sec:mmlm}

The goal of multilingual masked language modeling (MMLM;~\citealt{bert}) is to recover the masked tokens from a randomly masked sequence.
For each input sequence of MMLM, we sample a text from the monolingual corpus for pre-training.
Let $(c_1, x_1)$ denote a monolingual text sequence, where $x_1$ is the masked token, and $c_1$ is the corresponding context.
Intuitively, we need to maximize their dependency (i.e., $I(c_1; x_1)$), so that the context representations are predictive for masked tokens~\cite{infoword}.

For the example pair $(c_1 , x_1)$, we construct a set $\mathcal{N}$ that contains $x_1$ and $|\mathcal{N}|-1$ negative samples drawn from a proposal distribution $q$.
According to the InfoNCE~\cite{infonce} lower bound, we have:
\begin{align}
\label{eq:mlm:mi}
&I(c_1; x_1) \nonumber \\
&\geqslant \underset{q(\mathcal{N})}{E} \left[ \log \frac{f_\vtheta(c_1,x_1)}{\sum_{x'\in \mathcal{N}} f_\vtheta(c_1, x')} \right] + \log|\mathcal{N}|
\end{align}
where $f_\vtheta$ is a function that scores whether the input $c_1$ and $x_1$ is a positive pair.

Given context $c_1$, MMLM learns to minimize the cross-entropy loss of the masked token $x_1$:
\begin{align}
\label{eq:mlm:loss}
\Ls_{\textsc{MMLM}} = -\log \frac{\exp(g_{\vtheta_{T}}(c_1)^\top g_{\vtheta_{E}}(x_1))}{\sum_{x'\in \mathcal{V}}\exp(g_{\vtheta_{T}}(c_1)^\top g_{\vtheta_{E}}(x'))}
\end{align}
where $\mathcal{V}$ is the vocabulary, $g_{\vtheta_{E}}$ is a look-up function that returns the token embeddings, $g_{\vtheta_{T}}$ is a Transformer that returns the final hidden vectors in position of $x_1$.
According to \eqform{eq:mlm:mi} and \eqform{eq:mlm:loss}, if $\mathcal{N} = \mathcal{V}$ and $f_\vtheta(c_1, x_1)=\exp(g_{\vtheta_{T}}(c_1)^\top g_{\vtheta_{E}}(x_1))$, we can find that MMLM maximizes a lower bound of $I(c_1; x_1)$.

Next, we explain why MMLM can implicitly learn cross-lingual representations.
Let $(c_2 , x_2)$ denote a MMLM instance that is in different language as $(c_1 , x_1)$.
Because the vocabulary, the position embedding, and special tokens are shared across languages, it is common to find anchor points~\cite{pires2019multilingual,elements:mbert} where $x_1 = x_2$ (such as subword, punctuation, and digit) or $I(x_1 , x_2)$ is positive (i.e., the representations are associated or isomorphic).
With the bridge effect of $\{x_1 , x_2\}$, MMLM obtains a v-structure dependency ``$c_1 \rightarrow \{x_1 , x_2\} \leftarrow c_2$'', which leads to a negative co-information (i.e., interaction information) $I(c_1;c_2;\{x_1 , x_2\})$~\cite{triple:mi}.
Specifically, the negative value of $I(c_1;c_2;\{x_1 , x_2\})$ indicates that the variable $\{x_1 , x_2\}$ enhances the correlation between $c_1$ and $c_2$~\cite{fano63}.

In summary, although MMLM learns to maximize $I(c_1 , x_1)$ and $I(c_2 , x_2)$ in each language, we argue that the task encourages the cross-lingual correlation of learned representations.
Notice that for the setting without word-piece overlap~\cite{crosslingual:transfer:of:monolingual,emerging:xlm:acl20,xlingual:mbert:iclr20}, we hypothesize that the information bottleneck principle~\cite{info:bottleneck} tends to transform the cross-lingual structural similarity into isomorphic representations, which has similar bridge effects as the anchor points. Then we can explain how the cross-lingual ability is spread out as above. We leave more discussions about the setting without word-piece overlap for future work.

\subsection{Translation Language Modeling}
\label{sec:tlm}

Similar to MMLM, the goal of translation language modeling (TLM;~\citealt{xlm}) is also to predict masked tokens, but the prediction is conditioned on the concatenation of a translation pair.
We try to explain how TLM pre-training enhances cross-lingual transfer from an information-theoretic perspective.

Let $c_1$ and $c_2$ denote a translation pair of sentences, and $x_1$ a masked token taken in $c_1$.
So $c_1$ and $x_1$ are in the same language, while $c_1$ and $c_2$ are in different ones.
Following the derivations of MMLM in Section~\ref{sec:mmlm}, the objective of TLM is maximizing the lower bound of mutual information $I(c_1,c_2 ; x_1)$.
By re-writing the above mutual information, we have:
\begin{align}
\label{eq:tlm}
I(c_1,c_2;x_1) = I(c_1; x_1) + I(c_2; x_1 | c_1)
\end{align}
The first term $I(c_1; x_1)$ corresponds to MMLM, which learns to use monolingual context.
In contrast, the second term $I(c_2; x_1 | c_1)$ indicates cross-lingual mutual information between $c_2$ and $x_1$ that is not included by $c_1$. 
In other words, $I(c_2; x_1 | c_1)$ encourages the model to predict masked tokens by using the context in a different language.
In conclusion, TLM learns to utilize the context in both languages, which implicitly improves the cross-lingual transferability of pre-trained models.

\subsection{Cross-Lingual Contrastive Learning}
\label{sec:xlco}

Inspired by the unified information-theoretic framework, we propose a new cross-lingual pre-training task, named as cross-lingual contrast (\xlco{}).
The goal of \xlco{} is to maximize mutual information between the representations of parallel sentences $c_1$ and $c_2$, i.e., $I(c_1, c_2)$.
Unlike maximizing token-sequence mutual information in MMLM and TLM, \xlco{} targets at cross-lingual sequence-level mutual information.

We describe how the task is derived as follows.
Using InfoNCE~\cite{infonce} as the lower bound, we have:
\begin{align}
\label{eq:xlco}
&I(c_1; c_2) \geqslant \underset{q(\mathcal{N})}{E} \left[ \log \frac{f_\vtheta(c_1,c_2)}{\sum_{c'\in \mathcal{N}}f_\vtheta(c_1, c')} \right] + \log|\mathcal{N}|
\end{align}
where $\mathcal{N}$ is a set that contains the positive pair $c_2$ and $|\mathcal{N}|-1$ negative samples.
In order to maximize the lower bound of $I(c_1; c_2)$, we need to design the function $f_\vtheta$ that measures the similarity between the input sentence and the proposal distribution $q(\mathcal{N})$.
Specifically, we use the following similarity function $f_\vtheta$:
\begin{align}
f_\vtheta(c_1,c_2)= \exp( g_{\vtheta}(c_1)^\top g_{\vtheta}(c_2))
\end{align}
where $g_{\vtheta}$ is the Transformer encoder that we are pre-training.
Following~\cite{bert}, a special token \sptk{CLS} is added to the input, whose hidden vector is used as the sequence representation.
Additionally, we use a linear projection head after the encoder in $g_{\vtheta}$.

\paragraph{Momentum Contrast}
Another design choice is how to construct $\mathcal{N}$.
As shown in \eqform{eq:xlco}, a large $|\mathcal{N}|$ improves the tightness of the lower bound, which has been proven to be critical for contrastive learning~\cite{simclr}.

In our work, we employ the momentum contrast~\cite{moco} to construct the set $\mathcal{N}$, where the previously encoded sentences are progressively reused as negative samples.
Specifically, we construct two encoders with the same architecture which are the query encoder $g_{\vtheta_{Q}}$ and the key encoder $g_{\vtheta_{K}}$.
The loss function of \xlco{} is:
\begin{align}
\Ls_{\xlco{}} = -\log \frac{\exp(g_{\vtheta_{Q}}(c_1)^\top g_{\vtheta_{K}}(c_2))}{\sum_{c'\in \mathcal{N}}\exp(g_{\vtheta_{Q}}(c_1)^\top g_{\vtheta_{K}}(c'))}
\end{align}

During training, the query encoder $g_{\vtheta_{Q}}$ encodes $c_1$ and is updated by backpropagation.
The key encoder $g_{\vtheta_{K}}$ encodes $\mathcal{N}$ and is learned with momentum update~\cite{moco} towards the query encoder.
The negative examples in $\mathcal{N}$ are organized as a queue, where a newly encoded example is added while the oldest one is popped from the queue.
We initialize the query encoder and the key encoder with the same parameters, and pre-fill the queue with a set of encoded examples until it reaches the desired size $|\mathcal{N}|$.
Notice that the size of the queue remains constant during training.

\paragraph{Mixup Contrast}
For each pair, we concatenate it with a randomly sampled translation pair from another parallel corpus.
For example, consider the pairs $\langle c_1, c_2 \rangle$ and $\langle d_1, d_2 \rangle$ sampled from two different parallel corpora.
The two pairs are concatenated in a random order, such as $\langle c_1 d_1, c_2 d_2 \rangle$, and $\langle c_1 d_2, d_1 c_2 \rangle$.
The data augmentation of mixup encourages pre-trained models to learn sentence boundaries and to distinguish the order of multilingual texts.

\paragraph{Contrast on Universal Layer}
As a pre-training task maximizing the lower bound of sequence-level mutual information, \xlco{} is usually jointly learned with token-sequence tasks, such as MMLM, and TLM.
In order to make \xlco{} more compatible with the other pretext tasks, we propose to conduct contrastive learning on the most universal (or transferable) layer in terms of MMLM and TLM.

In our implementations, we instead use the hidden vectors of \sptk{CLS} at layer 8 to perform contrastive learning for base-size (12 layers) models, and layer 12 for large-size (24 layers) models.
Because previous analysis~\cite{sabet2020simalign,elements:mbert,emerging:xlm:acl20} shows that the specific layers of MMLM learn more universal representations and work better on cross-lingual retrieval tasks than other layers.
We choose the layers following the same principle.

The intuition behind the method is that MMLM and TLM encourage the last layer to produce language-distinguishable token representations because of the masked token classification.
But \xlco{} tends to learn similar representations across languages.
So we do not directly use the hidden states of the last layer in \xlco{}.

\subsection{Cross-Lingual Pre-Training}

% loss
We pretrain a cross-lingual model \our{} by jointly maximizing the lower bounds of three types of mutual information, including monolingual token-sequence mutual information (MMLM), cross-lingual token-sequence mutual information (TLM), and cross-lingual sequence-level mutual information (\xlco{}).
Formally, the loss of cross-lingual pre-training in \our{} is defined as:
\begin{align}
\Ls = \Ls_{\textsc{MMLM}} + \Ls_{\textsc{TLM}} + \Ls_{\text{\xlco{}}}
\end{align}
where we apply the same weight for the loss terms.

Both TLM and \xlco{} use parallel data.
The number of bilingual pairs increases with the square of the number of languages.
In our work, we set English as the pivot language following~\cite{xlm}, i.e., we only use the parallel corpora that contain English.

In order to balance the data size between high-resource and low-resource languages, we apply a multilingual sampling strategy~\cite{xlm} for both monolingual and parallel data.
An example in the language $l$ is sampled with the probability $p_{l} \propto (n_{l}/n)^{0.7}$, where $n_{l}$ is the number of instances in the language $l$, and $n$ refers to the total number of data.
Empirically, the sampling algorithm alleviates the bias towards high-resource languages~\cite{xlmr}.

\section{Experiments}
\label{sec:exp}

In this section, we first present the training configuration of \our{}.
Then we compare the fine-tuning results of \our{} with previous work on three cross-lingual understanding tasks.
We also conduct ablation studies to understand the major components of \our{}.

\begin{table*}[t]
\centering
\scalebox{0.78}{
% \scriptsize
\renewcommand\tabcolsep{3.9pt}
\begin{tabular}{lccccccccccccccccc}
\toprule
Models & \#M & en & fr & es & de & el & bg & ru & tr & ar & vi & th & zh & hi & sw & ur & Avg \\ \midrule
\multicolumn{18}{l}{\textit{Fine-tune multilingual model on English training set (Cross-lingual Transfer)}} \\
\midrule
\textsc{mBert}* & N & 82.1 & 73.8 & 74.3 & 71.1 & 66.4 & 68.9 & 69.0 & 61.6 & 64.9 & 69.5 & 55.8 & 69.3 & 60.0 & 50.4 & 58.0 & 66.3 \\
\textsc{XLM} (w/o TLM)* & N & 83.7 & 76.2 & 76.6 & 73.7 & 72.4 & 73.0 & 72.1 & 68.1 & 68.4 & 72.0 & 68.2 & 71.5 & 64.5 & 58.0 & 62.4 & 71.3 \\
\textsc{XLM}* & N & 85.0 & 78.7 & 78.9 & 77.8 & 76.6 & 77.4 & 75.3 & 72.5 & 73.1 & 76.1 & 73.2 & 76.5 & 69.6 & 68.4 & 67.3 & 75.1 \\
\textsc{XLM} (w/o TLM)* & 1 & 83.2 & 76.7 & 77.7 & 74.0 & 72.7 & 74.1 & 72.7 & 68.7 & 68.6 & 72.9 & 68.9 & 72.5 & 65.6 & 58.2 & 62.4 & 70.7 \\
\textsc{Unicoder} & 1 & 85.4 & 79.2 & 79.8 & 78.2 & 77.3 & 78.5 & 76.7 & 73.8 & 73.9 & 75.9 & 71.8 & 74.7 & 70.1 &  67.4 & 66.3 & 75.3 \\
\textsc{XLM-R}* & 1 & 85.8 & 79.7 & 80.7 & 78.7 & 77.5 & 79.6 & 78.1 & 74.2 & 73.8 & 76.5 & 74.6 & 76.7 & 72.4 & 66.5 & 68.3 & 76.2 \\
\textsc{XLM-R} (reimpl) & 1 & 84.7 & 79.1 & 79.4 & 77.4 & 76.6 & 78.4 & 76.0 & 73.5 & 72.6 & 75.5 & 73.0 & 74.5 & 71.0 & 65.7 & 67.6 & 75.0 \\
\our{} & 1 & 86.4 & 80.3 & 80.9 & 79.3 & 77.8 & 79.3 & 77.6 & 75.6 & 74.2 & 77.1 & 74.6 & 77.0 & 72.2 & 67.5 & 67.3 & \textbf{76.5} \\
~~$-$\xlco{} & 1 & 86.5 & 80.5 & 80.3 & 78.7 & 77.3 & 78.8 & 77.4 & 74.6 & 73.8 & 76.8 & 73.7 & 76.7 & 71.8 & 66.3 & 66.4 & 76.0 \\
\midrule
\textsc{XLM-R}$_\textsc{LARGE}$* & 1 & 89.1 & 84.1 & 85.1 & 83.9 & 82.9 & 84.0 & 81.2 & 79.6 & 79.8 & 80.8 & 78.1 & 80.2 & 76.9 & 73.9 & 73.8 & 80.9  \\
\textsc{XLM-R}$_\textsc{LARGE}$ (reimpl) & 1 & 88.9 & 83.6 & 84.8 & 83.1 & 82.4 & 83.7 & 80.7 & 79.2 & 79.0 & 80.4 & 77.8 & 79.8 & 76.8 & 72.7 & 73.3 & 80.4 \\
\our{}$_\textsc{LARGE}$ & 1 & 89.7 & 84.5 & 85.5 & 84.1 & 83.4 & 84.2 & 81.3 & 80.9 & 80.4 & 80.8 & 78.9 & 80.9 & 77.9 & 74.8 & 73.7 & \textbf{81.4} \\
\midrule
\multicolumn{18}{l}{\textit{Fine-tune multilingual model on all training sets (Translate-Train-All)}} \\
\midrule
\textsc{XLM} (w/o TLM)* & 1 & 84.5 & 80.1 & 81.3 & 79.3 & 78.6 & 79.4 & 77.5 & 75.2 & 75.6 & 78.3 & 75.7 & 78.3 & 72.1 & 69.2 & 67.7 & 76.9 \\
\textsc{XLM}* & 1 & 85.0 & 80.8 & 81.3 & 80.3 & 79.1 & 80.9 & 78.3 & 75.6 & 77.6 & 78.5 & 76.0 & 79.5 & 72.9 & 72.8 & 68.5 & 77.8 \\
\textsc{XLM-R}* & 1 & 85.4 & 81.4 & 82.2 & 80.3 & 80.4 & 81.3 & 79.7 & 78.6 & 77.3 & 79.7 & 77.9 & 80.2 & 76.1 & 73.1 & 73.0 & 79.1 \\
\textsc{XLM-R} (reimpl) & 1 & 85.0 & 81.0 & 81.9 & 80.6 & 79.7 & 81.4 & 79.5 & 77.7 & 77.3 & 79.5 & 77.5 & 79.1 & 75.3 & 72.2 & 70.9 & 78.6 \\
\our{} & 1 & 86.5 & 82.6 & 83.0 & 82.3 & 81.3 & 82.4 & 80.6 & 79.5 & 78.9 & 81.0 & 78.9 & 80.7 & 77.8 & 73.3 & 71.6 & \textbf{80.0} \\
\bottomrule
\end{tabular}
}
\caption{Evaluation results on XNLI cross-lingual natural language inference.
We report test accuracy in $15$ languages.
The model number \#M=N indicates the model selection is done on each language's validation set (i.e., each language has a  different model), while \#M=1 means only one model is used for all languages.
Results with ``*'' are taken from~\citet{xlmr}.
``(reimpl)'' is our reimplementation of fine-tuning, which is the same as \our{}.
Results of \our{} and XLM-R (reimpl) are averaged over five runs.
``$-$\xlco{}'' is the model without cross-lingual contrast.}
\label{table:xnli}
\end{table*}

\subsection{Setup}

\paragraph{Corpus}
We use the same pre-training corpora as previous models~\cite{xlmr,xlm}.
Specifically, we reconstruct CC-100~\cite{xlmr} for MMLM, which remains $94$ languages by filtering the language code larger than 0.1GB.
Following~\cite{xlm}, for the TLM and \xlco{} tasks, we employ $14$ language pairs of parallel data that involves English.
We collect translation pairs from MultiUN~\cite{multiun}, IIT Bombay~\cite{iit}, OPUS~\cite{opus}, and WikiMatrix~\cite{wikimatrix}.
The size of parallel corpora is about 42GB.
More details about the pre-training data are described in the appendix.

\paragraph{Model Size}
We follow the model configurations of XLM-R~\cite{xlmr}.
For the Transformer~\cite{transformer} architecture, we use 12 layers and 768 hidden states for \our{} (i.e., base size), and 24 layers and 1,024 hidden states for \our{}$_\textsc{LARGE}$ (i.e., large size).

\paragraph{Hyperparameters}
We initialize the parameters of \our{} with XLM-R.
We optimize the model with Adam~\cite{adam} using a batch size of $2048$ for a total of 150K steps for \our{}, and 200K steps for \our{}$_\textsc{LARGE}$.
The same number of training examples are fed to three tasks.
The learning rate is scheduled with a linear decay with 10K warmup steps, where the peak learning rate is set as $0.0002$ for \our{}, and $0.0001$ for \our{}$_\textsc{LARGE}$.
The momentum coefficient is set as $0.9999$ and $0.999$ for \our{} and \our{}$_\textsc{LARGE}$, respectively. The length of the queue is set as $131,072$.
The training procedure takes about $2.3$ days $\times$ $2$ Nvidia DGX-2 stations for \our{}, and $5$ days $\times$ $16$ Nvidia DGX-2 stations for \our{}$_\textsc{LARGE}$.
Details about the pre-training hyperparameters can be found in the appendix.

\subsection{Evaluation}

We conduct experiments over three cross-lingual understanding tasks, i.e., cross-lingual natural language inference, cross-lingual sentence retrieval, and cross-lingual question answering.

\paragraph{Cross-Lingual Natural Language Inference}
The Cross-Lingual Natural Language Inference corpus (XNLI;~\citealt{xnli}) is a widely used cross-lingual classification benchmark. The goal of NLI is to identify the relationship of an input sentence pair. 
We evaluate the models under the following two settings.
(1) Cross-Lingual Transfer: fine-tuning the model with English training set and directly evaluating on multilingual test sets.
(2) Translate-Train-All: fine-tuning the model with the English training data and the pseudo data that are translated from English to the other languages.

\paragraph{Cross-Lingual Sentence Retrieval}
The goal of the cross-lingual sentence retrieval task is to extract parallel sentences from bilingual comparable corpora. 
We use the subset of 36 language pairs of the Tatoeba dataset~\cite{tatoeba} for the task.
The dataset is collected from Tatoeba\footnote{\url{https://tatoeba.org/eng/}},
which is an open collection of multilingual parallel sentences in more than 300 languages.
Following~\cite{xtreme}, we use the averaged hidden vectors in the seventh Transformer layer to compute cosine similarity for sentence retrieval.

\paragraph{Cross-Lingual Question Answering}
We use the Multilingual Question Answering (MLQA;~\citealt{mlqa}) dataset for the cross-lingual QA task. MLQA provides development and test data in seven languages in the format of SQuAD v1.1~\cite{squad1}.
We follow the fine-tuning method introduced in~\cite{bert} that concatenates the question-passage pair as the input.

\subsection{Results}
\label{sec:results}

We compare \our{} with the following pre-trained Transformer models:
(1) Multilingual BERT (\textsc{mBert};~\citealt{bert}) is pre-trained with MMLM on Wikipedia in $102$ languages;
(2) \textsc{XLM}~\cite{xlm} pretrains both MMLM and TLM tasks on Wikipedia in $100$ languages;
(3) \textsc{XLM-R}~\cite{xlmr} scales up MMLM to the large CC-100 corpus in $100$ languages with much more training steps;
(4) \textsc{Unicoder}~\cite{xglue} continues training \textsc{XLM-R} with MMLM and TLM.
(5) \our{}$-$\xlco{} continues training \textsc{XLM-R} with MMLM and TLM, using the same pre-training datasets with \our{}.

\paragraph{Cross-Lingual Natural Language Inference}
Table~\ref{table:xnli} reports the classification accuracy on each test of XNLI under the above evaluation settings.
The final scores on test set are averaged over five random seeds.
\our{} outperforms all baseline models on the two evaluation settings of XNLI.
In the cross-lingual transfer setting, \our{} achieves 76.5 averaged accuracy, outperforming XLM-R (reimpl) by 1.5.
Similar improvements can be observed for large-size models.
Moreover, the ablation results ``$-$\xlco{}'' show that cross-lingual contrast is helpful for zero-shot transfer in most languages.
We also find that \our{} improves the results in the translate-train-all setting.

\begin{table*}[t]
\centering
\scalebox{0.86}{
\renewcommand\tabcolsep{3.8pt}
\begin{tabular}{lccccccccccccccccc}
\toprule
Models & Direction & ar & bg & zh & de & el & fr & hi & ru & es & sw & th & tr & ur & vi & Avg \\ \midrule
\textsc{XLM-R} & xx $\rightarrow$ en  & 36.8  & 67.6  & 60.7  & 89.9  & 53.7  & 74.1  & 54.2  & 72.5  & 74.0  & 18.7  & 38.3  & 61.1  & 36.6  & 68.4  & 57.6 \\
\our{} & xx $\rightarrow$ en  & 59.0  & 78.6  & 86.3  & 93.9  & 62.1  & 79.4  & 87.1  & 83.8  & 88.2  & 39.5  & 84.9  & 83.3  & 73.0  & 89.6  & \textbf{77.8} \\ 
~~$-$\xlco{} & xx $\rightarrow$ en  & 42.9  & 65.5  & 69.5  & 91.1  & 55.6  & 76.4  & 71.6  & 74.9  & 74.8  & 20.5  & 68.1  & 69.8  & 51.6  & 81.8  & 65.3 \\
\midrule
\textsc{XLM-R} & en $\rightarrow$ xx  & 38.6  & 69.9  & 60.3  & 89.4  & 57.3  & 74.3  & 49.3  & 73.0  & 74.6  & 14.4  & 58.4  & 64.0  & 36.9  & 72.5  & 59.5 \\
\our{} & en $\rightarrow$ xx  & 68.6  & 78.6  & 86.4  & 95.1  & 72.6  & 84.0  & 88.3  & 85.7  & 87.2  & 40.8  & 91.2  & 84.7  & 73.3  & 92.0  & \textbf{80.6} \\
~~$-$\xlco{} & en $\rightarrow$ xx  & 45.4  & 64.0  & 69.3  & 88.1  & 56.5  & 72.3  & 69.6  & 73.6  & 71.5  & 22.1  & 79.7  & 64.3  & 48.2  & 79.8  & 64.6 \\
\bottomrule
\end{tabular}
}
\caption{Evaluation results on Tatoeba cross-lingual sentence retrieval. We report the top-1 accuracy of $14$ language pairs that are covered by parallel data.}
\label{table:xir15}
\end{table*}

\begin{table*}[t]
\centering
\scriptsize
\renewcommand\tabcolsep{2.3pt}
\begin{tabular}{lccccccccccccccccccccccccccc}
\toprule
Models & Direction & af & bn & et & eu & fi & he & hu & id & it & jv & ja & ka & kk & ko & ml & mr & nl & fa & pt & ta & te & tl & Avg \\ \midrule
\textsc{XLM-R} & xx $\rightarrow$ en  & 55.2  & 29.3  & 49.3  & 33.5  & 66.7  & 53.9  & 61.6  & 70.8  & 68.2  & 15.1  & 57.2 & 41.4  & 40.3  & 51.6  & 56.5  & 46.0  & 79.5  & 68.0  & 80.6  & 25.7  & 32.5  & 31.2  & 50.6 \\
\our{} & xx $\rightarrow$ en  & 48.6  & 49.6  & 38.3  & 36.7  & 65.7  & 62.9  & 61.7  & 79.9  & 72.2  & 13.2  & 78.3 & 57.4  & 49.2  & 74.5  & 76.6  & 72.0  & 80.8  & 82.2  & 84.7  & 53.7  & 53.0  & 42.1  & \textbf{60.6} \\
~~$-$\xlco{} & xx $\rightarrow$ en  & 33.1  & 33.5  & 25.9  & 20.8  & 48.4  & 49.1  & 46.1  & 68.5  & 60.4  & 12.2  & 60.6 & 38.6  & 35.1  & 60.6  & 57.8  & 49.1  & 72.2  & 66.0  & 75.3  & 36.5  & 38.0  & 25.5  & 46.1 \\
\midrule
\textsc{XLM-R} & en $\rightarrow$ xx & 55.0  & 27.9  & 50.2  & 32.5  & 72.9  & 63.2  & 67.1  & 71.9  & 68.0   & 9.8  & 58.2 & 52.0  & 41.7  & 58.3  & 60.8  & 42.1  & 78.9  & 69.6  & 82.1  & 33.2  & 38.9  & 29.7  & 52.9 \\
\our{} & en $\rightarrow$ xx & 51.8  & 49.1  & 35.2  & 28.6  & 65.6  & 66.5  & 61.7  & 80.1  & 72.8   & 7.8  & 80.4 & 61.9  & 50.6  & 79.6  & 78.7  & 68.1  & 81.8  & 82.8  & 86.5  & 63.5  & 53.0  & 35.5  & \textbf{61.0} \\
~~$-$\xlco{} & en $\rightarrow$ xx & 28.1  & 23.5  & 19.0  & 12.6  & 45.2  & 49.7  & 40.8  & 62.8  & 57.5   & 3.4  & 58.2 & 38.9  & 31.3  & 61.0  & 57.5  & 37.2  & 67.8  & 66.4  & 75.0  & 43.0  & 31.6  & 17.9  & 42.2 \\
\bottomrule
\end{tabular}
\caption{Evaluation results on Tatoeba cross-lingual sentence retrieval. We report the top-1 accuracy scores of $22$ language pairs that are not covered by parallel data.}
\label{table:xirno15}
\end{table*}

\paragraph{Cross-Lingual Sentence Retrieval}
% results
In Table~\ref{table:xir15} and Table~\ref{table:xirno15}, we report the top-1 accuracy scores of cross-lingual sentence retrieval with the base-size models.
The evaluation results demonstrate that \our{} produces better aligned cross-lingual sentence representations.
On the 14 language pairs that are covered by parallel data, \our{} obtains 77.8 and 80.6 averaged top-1 accuracies in the directions of xx $\rightarrow$ en and en $\rightarrow$ xx, outperforming XLM-R by 20.2 and 21.1.
Even on the 22 language pairs that are not covered by parallel data, \our{} outperforms XLM-R on 16 out of 22 language pairs, providing 8.1\% improvement in averaged accuracy.
In comparison, the ablation variant ``$-$\xlco{}'' (i.e., MMLM$+$TLM) obtains better results than XLM-R in Table~\ref{table:xir15}, while getting worse performance than XLM-R in Table~\ref{table:xirno15}.
The results indicate that \xlco{} encourages the model to learn universal representations even on the language pairs without parallel supervision.

\begin{table*}[t]
\centering
\scalebox{0.76}{
\renewcommand\tabcolsep{5.5pt}
\begin{tabular}{lccccccccc}
\toprule
Models & en & es & de & ar & hi & vi & zh & Avg \\ \midrule
\textsc{mBert}* & 77.7 / 65.2 & 64.3 / 46.6 & 57.9 / 44.3 & 45.7 / 29.8 & 43.8 / 29.7 & 57.1 / 38.6 & 57.5 / 37.3 & 57.7 / 41.6 \\
XLM* & 74.9 / 62.4 & 68.0 / 49.8 & 62.2 / 47.6 & 54.8 / 36.3 & 48.8 / 27.3 & 61.4 / 41.8 & 61.1 / 39.6 & 61.6 / 43.5 \\
\textsc{Unicoder} & 80.6 / ~~~-~~~ &  68.6 / ~~~-~~~ &  62.7 / ~~~-~~~ & 57.8 / ~~~-~~~ & 62.7 / ~~~-~~~ &  67.5 / ~~~-~~~ & 62.1 / ~~~-~~~ & 66.0 / ~~~-~~~ \\
XLM-R &  77.1 / 64.6 & 67.4 / 49.6 & 60.9 / 46.7 & 54.9 / 36.6 & 59.4 / 42.9 & 64.5 / 44.7 & 61.8 / 39.3 & 63.7 / 46.3 \\
XLM-R (reimpl) & 80.2 / 67.0 & 67.7 / 49.9 & 62.1 / 47.7 & 56.1 / 37.2 & 61.1 / 44.0 & 67.0 / 46.3 & 61.4 / 38.5 & 65.1 / 47.2 \\
\our{} & 81.6 / 68.3 & 69.8 / 51.6 & 64.3 / 49.4 & 60.6 / 40.9 & 65.2 / 47.1 & 70.2 / 49.0 & 64.8 / 41.3 & \textbf{68.1} / \textbf{49.6} \\
~~$-$\xlco{} & 81.2 / 68.1 & 69.6 / 51.9 & 64.0 / 49.3 & 59.7 / 40.2 & 64.0 / 46.3 & 69.3 / 48.0 & 64.1 / 40.6 & 67.4 / 49.2 \\
\midrule
XLM-R$_\textsc{LARGE}$ & 80.6 / 67.8 & 74.1 / 56.0 & 68.5 / 53.6 & 63.1 / 43.5 & 69.2 / 51.6 & 71.3 / 50.9 & 68.0 / 45.4 & 70.7 / 52.7 \\
XLM-R$_\textsc{LARGE}$ (reimpl) & 84.0 / 71.1 & 74.4 / 56.4 & 70.2 / 55.0 & 66.5 / 46.3 & 71.1 / 53.2 & 74.4 / 53.5 & 68.6 / 44.6 & 72.7 / 54.3 \\
\our{}$_\textsc{LARGE}$ & 84.5 / 71.6 & 75.1 / 57.3 & 71.2 / 56.2 & 67.6 / 47.6 & 72.5 / 54.2 & 75.2 / 54.1 & 69.2 / 45.4 & \textbf{73.6} / \textbf{55.2} \\
\bottomrule
\end{tabular}
}
\caption{Evaluation results on MLQA cross-lingual question answering. We report the F1 and exact match (EM) scores.
Results with ``*'' are taken from \cite{mlqa}.
``(reimpl)'' is our reimplementation of fine-tuning, which is the same as \our{}.
Results of \our{} and XLM-R (reimpl) are averaged over five runs.
``$-$\xlco{}'' is the model without cross-lingual contrast.
}
\label{table:mlqa}
\end{table*}

\paragraph{Cross-Lingual Question Answering}
Table~\ref{table:mlqa} compares \our{} with baseline models on MLQA, where we report the F1 and the exact match (EM) scores on each test set.
Both \our{} and \our{}$_\textsc{LARGE}$ obtain the best results against the four baselines.
In addition, the results of the ablation variant ``$-$\xlco{}'' indicate that the proposed cross-lingual contrast is beneficial on MLQA.

\subsection{Analysis and Discussion}

To understand \our{} and the cross-lingual contrast task more deeply, 
we conduct analysis from the perspectives of cross-lingual transfer and cross-lingual representations.
Furthermore, we perform comprehensive ablation studies on the major components of \our{}, including the cross-lingual pre-training tasks, mixup contrast, the contrast layer, and the momentum contrast. To reduce the computation load, we use \our{}15 in our ablation studies, which is trained on 15 languages for 100K steps.

\begin{table}[t]
\centering
\scalebox{0.88}{
\begin{tabular}{lcccc}
\toprule
Models & MLQA & XNLI & Average \\ \midrule
\textsc{mBert} & 23.3 & 16.9 & 20.1 \\
XLM-R  & 17.6 & 10.4 & 14.0 \\ 
\our{} & \textbf{15.8} & \textbf{10.3} & \textbf{13.1} \\
~~$-$\xlco{} & 16.1 & 11.0 & 13.6 \\
\bottomrule
\end{tabular}
}
\caption{Cross-lingual transfer gap scores, i.e., averaged performance drop between English and other languages in zero-shot transfer. Smaller gap indicates better transferability.
``$-$\xlco{}'' is the model without cross-lingual contrast.
}
\label{table:gap}
\end{table}

\paragraph{Cross-Lingual Transfer Gap}
Cross-lingual transfer gap~\cite{xtreme} is the difference between the performance on the English test set and the averaged performance on the test sets of all other languages. A lower cross-lingual transfer gap score indicates more end-task knowledge from the English training set is transferred to other languages. 
In Table~\ref{table:gap}, we compare the cross-lingual transfer gap scores of \our{} with baseline models on MLQA and XNLI. 
Note that we do not include the results of XLM because it is pre-trained on 15 languages or using \#M=N.
The results show that \our{} reduces the gap scores on both MLQA and XNLI, providing better cross-lingual transferability than the baselines.

\begin{figure}[t]
\begin{center} 
\includegraphics[width=0.96\linewidth]{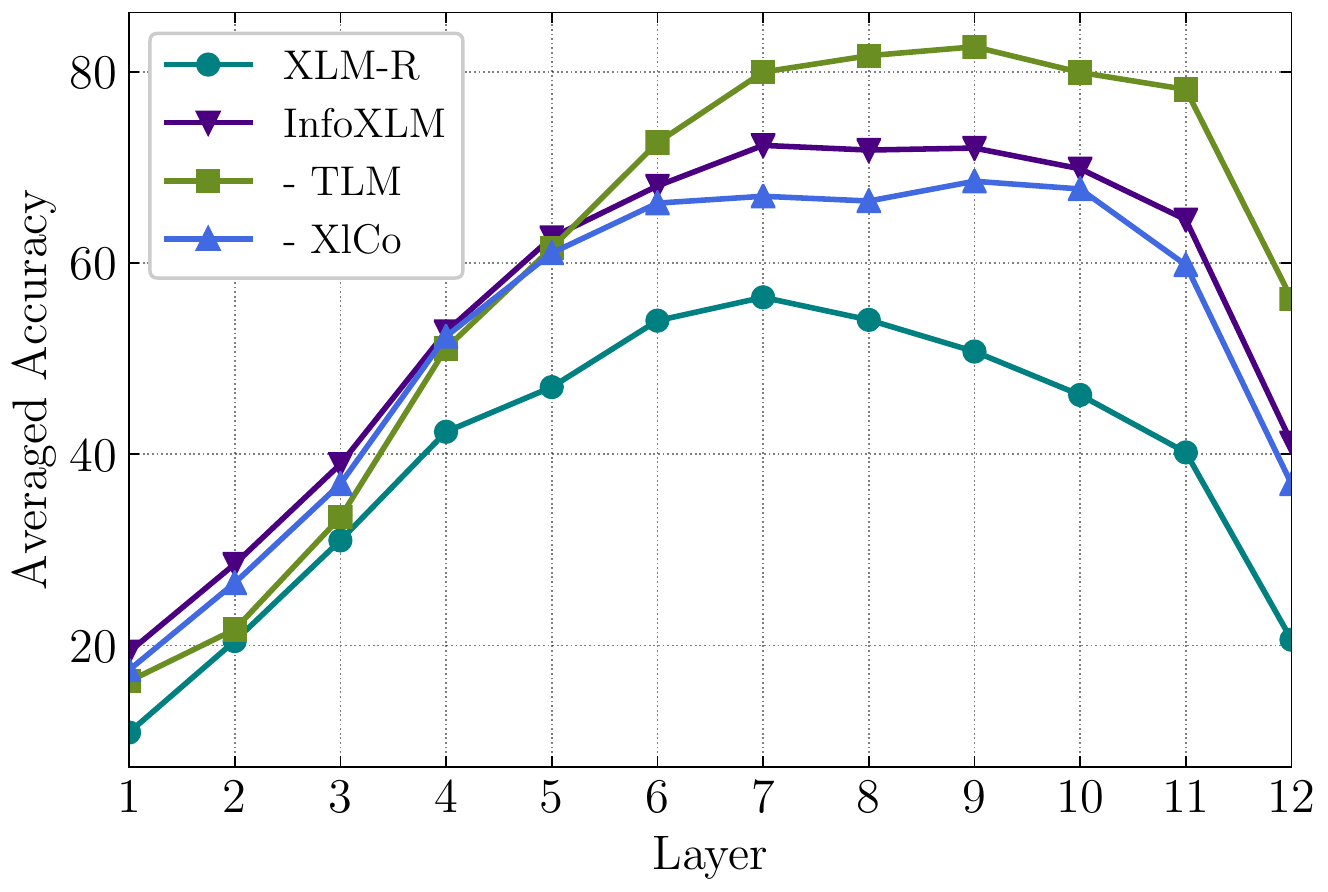}
\caption{Evaluation results of different layers on Tatoeba cross-lingual sentence retrieval.}
\label{fig:layerrep}
\end{center} 
\end{figure}

\paragraph{Cross-Lingual Representations}

In addition to cross-lingual transfer, learning good cross-lingual representations is also the goal of cross-lingual pre-training. 
In order to analyze how the cross-lingual contrast task affects the alignment of the learned cross-lingual representations,  we evaluate the representations of different middle layers on the Tatoeba test sets of the 14 languages that are covered by parallel data.
Figure \ref{fig:layerrep} presents the averaged top-1 accuracy of cross-lingual sentence retrieval in the direction of xx $\rightarrow$ en. 
\our{} outperforms \textsc{XLM-R} on all of the 12 layers, demonstrating that our proposed task improves the cross-lingual alignment of the learned representations.
From the results of XLM-R, we observe that the model suffers from a performance drop in the last few layers.
The reason is that MMLM encourages the representations of the last hidden layer to be similar to token embeddings, which is contradictory with the goal of learning cross-lingual representations.
In contrast, \our{} still provides high retrieval accuracy at the last few layers, which indicates that \our{} provides better aligned representations than XLM-R.
Moreover, we find that the performance is further improved when removing TLM, demonstrating that \xlco{} is more effective than TLM for aligning cross-lingual representations, although TLM helps to improve zero-shot cross-lingual transfer.

\paragraph{Effect of Cross-Lingual Pre-training Tasks}
To better understand the effect of the cross-lingual pre-training tasks, we perform ablation studies on the pre-training tasks of \our{}, by removing \xlco{}, TLM, or both.
We present the experimental results in Table~\ref{table:ablations}.
Comparing the results of $-$TLM and $-$\xlco{} with the results of $-$TLM$-$\xlco{},
we find that both \xlco{} and TLM effectively improve cross-lingual transferability of the pre-trained \our{} model. TLM is more effective for XNLI while \xlco{} is more effective for MLQA. Moreover, the performance can be further improved by jointly learning \xlco{} and TLM.

\begin{table}[t]
\centering
% \scriptsize
\scalebox{0.78}{
% \small
\renewcommand\tabcolsep{4.5pt}
\begin{tabular}{lccccc}
\toprule
Model & \xlco{}{} Layer & XNLI & MLQA \\ \midrule
\our{}15 & 8 & 76.45 & 67.87 / 49.58 \\
\our{}15 & 12 & 76.12 & 67.83 / 49.50 \\
\midrule
\our{}15$-$\textsc{TLM} & 8 & 75.58 & 67.42 / 49.27 \\ 
\our{}15$-$\textsc{TLM} & 12 & 75.85 & 67.84 / 49.54 \\
\bottomrule
\end{tabular}
}
\caption{Contrast on the universal layer v.s. on the last layer. Results are averaged over five runs.
``$-$\textsc{TLM}'' is the ablation variant without TLM.
}
\label{table:coxl-layer}
\end{table}

\paragraph{Effect of Contrast on Universal Layer}
We conduct experiments to investigate whether contrast on the universal layer improves cross-lingual pre-training.
As shown in Table~\ref{table:coxl-layer}, we compare the evaluation results of four variants of \our{}, where \xlco{} is applied on the layer 8 (i.e., universal layer) or on the layer 12 (i.e., the last layer).
We find that contrast on the layer 8 provides better results for \our{}. 
However, conducting \xlco{} on layer 12 performs better when the TLM task is excluded.
The results show that maximizing context-sequence (TLM) and sequence-level (\xlco{}) mutual information at the last layer tends to interfere with each other.
Thus, we suggest applying \xlco{} on the universal layer for pre-training \our{}.

\begin{table}[t]
\centering
\scalebox{0.84}{
\begin{tabular}{llccccc}
\toprule
&Model & XNLI & MLQA \\ \midrule
\tblidx{0} &\our{}15 & \textbf{76.45} & \textbf{67.87} / \textbf{49.58} \\
\tblidx{1} &\tblidx{0}$-$\xlco{} & 76.24 & 67.43 / 49.23 \\
\tblidx{2} &\tblidx{0}$-$TLM & 75.85 & 67.84 / 49.54 \\
\tblidx{3} &\tblidx{2}$-$\xlco{} & 75.33 & 66.86 / 48.82 \\ 
\tblidx{4} &\tblidx{2}$-$Mixup & 75.43 & 67.21 / 49.19 \\
\tblidx{5} &\tblidx{2}$-$Momentum & 75.32 & 66.58 / 48.66 \\
\bottomrule
\end{tabular}
}
\caption{Ablation results on components of \our{}. Results are averaged over five runs.}
% \vspace{-5mm}
\label{table:ablations}
\end{table}

\paragraph{Effect of Mixup Contrast}
We conduct an ablation study on the mixup contrast strategy.
We pretrain a model that directly uses translation pairs for \xlco{} without mixup contrast ($-$TLM$-$Mixup).
As shown in Table~\ref{table:ablations}, we present the evaluation results on XNLI and MLQA.
We observe that mixup contrast improves the performance of \our{} on both datasets.

\paragraph{Effect of Momentum Contrast}
In order to show whether our pre-trained model benefits from momentum contrast, we pretrain a revised version of \our{} without momentum contrast.
In other words, the parameters of the key encoder are always the same as the query encoder.
As shown in Table~\ref{table:ablations}, we report evaluation results (indicated by ``$-$TLM$-$Momentum'') of removing momentum contrast on XNLI and MLQA.
We observe a performance descent after removing the momentum contrast from \our{}, which indicates that momentum contrast improves the learned language representations of \our{}.

\section{Conclusion}

In this paper, we present a cross-lingual pre-trained model \our{} that is trained with both monolingual and parallel corpora.
The model is motivated by the unified view of cross-lingual pre-training from an information-theoretic perspective.
Specifically, in addition to the masked language modeling and translation language modeling tasks, \our{} is jointly pre-trained with a newly introduced cross-lingual contrastive learning task.
The cross-lingual contrast leverages bilingual pairs as the two views of the same meaning, and encourages their encoded representations to be more similar than the negative examples.
Experimental results on several cross-lingual language understanding tasks show that~\our{} can considerably improve the performance.

\section{Ethical Considerations}

Currently, most NLP research works and applications are English-centric, which makes non-English users hard to access to NLP-related services. Our work focuses on cross-lingual language model pre-training. With the pre-trained model, we are able to transfer end-task knowledge from high-resource languages to low-resource languages, which helps to build more accessible NLP applications.
Additionally, incorporating parallel corpora into the pre-training procedure improves the training efficiency, which potentially reduces the computational cost for building multilingual NLP applications.

\section*{Acknowledgements}
We appreciate the helpful discussions with Bo Zheng, Shaohan Huang, Shuming Ma, and Yue Cao.

\bibliographystyle{acl_natbib}
\bibliography{cxlm}

\newpage
\appendix

\section{Pre-Training Data}

We reconstruct CCNet\footnote{\url{https://github.com/facebookresearch/cc_net}} and follow~\cite{xlmr} to reproduce the CC-100 corpus for monolingual texts.
The resulting corpus contains $94$ languages.
Table~\ref{table:cc} reports the language codes and data size in our work.
Notice that several languages can share the same ISO language code, e.g., zh represents both Simplified Chinese and Traditional Chinese.
Moreover, Table~\ref{table:parallel:data} shows the statistics of the parallel data.

\begin{table}[ht]
\centering
\scriptsize
\begin{tabular}{crcrcr}
\toprule
Code & Size (GB) & Code & Size (GB) & Code & Size (GB) \\ \cmidrule(r){1-2}\cmidrule{3-4}\cmidrule(l){5-6}
af & 0.2 & hr & 1.4 & pa & 0.8 \\
am & 0.4 & hu & 9.5 & pl & 28.6 \\
ar & 16.1 & hy & 0.7 & ps & 0.4 \\
as & 0.1 & id & 17.2 & pt & 39.4 \\
az & 0.8 & is & 0.5 & ro & 11.0 \\
ba & 0.2 & it & 47.2 & ru & 253.3 \\
be & 0.5 & ja & 86.8 & sa & 0.2 \\
bg & 7.0 & ka & 1.0 & sd & 0.2 \\
bn & 5.5 & kk & 0.6 & si & 1.3 \\
ca & 3.0 & km & 0.2 & sk & 13.6 \\
ckb & 0.6 & kn & 0.3 & sl & 6.2 \\
cs & 14.9 & ko & 40.0 & sq & 3.0 \\
cy & 0.4 & ky & 0.5 & sr & 7.2 \\
da & 6.9 & la & 0.3 & sv & 60.4 \\
de & 99.0 & lo & 0.2 & sw & 0.3 \\
el & 13.1 & lt & 2.3 & ta & 7.9 \\
en & 731.6 & lv & 1.3 & te & 2.3 \\
eo & 0.5 & mk & 0.6 & tg & 0.7 \\
es & 85.6 & ml & 1.3 & th & 33.0 \\
et & 1.4 & mn & 0.4 & tl & 1.2 \\
eu & 1.0 & mr & 0.5 & tr & 56.4 \\
fa & 19.0 & ms & 0.7 & tt & 0.6 \\
fi & 5.9 & mt & 0.2 & ug & 0.2 \\
fr & 89.9 & my & 0.4 & uk & 13.4 \\
ga & 0.2 & ne & 0.6 & ur & 3.0 \\
gl & 1.5 & nl & 25.9 & uz & 0.1 \\
gu & 0.3 & nn & 0.4 & vi & 74.5 \\
he & 4.4 & no & 5.5 & yi & 0.3 \\
hi & 5.0 & or & 0.3 & zh & 96.8 \\
\bottomrule
\end{tabular}
\caption{The statistics of CCNet used corpus for pre-training.}
\label{table:cc}
\end{table}

\begin{table}[ht]
\centering
\scriptsize
\begin{tabular}{crcr}
\toprule
ISO Code & Size (GB) & ISO Code & Size (GB) \\ \midrule
en-ar & 5.88 & en-ru & 7.72 \\
en-bg & 0.49 & en-sw & 0.06 \\
en-de & 4.21 & en-th & 0.47 \\
en-el & 2.28 & en-tr & 0.34 \\
en-es & 7.09 & en-ur & 0.39 \\
en-fr & 7.63 & en-vi & 0.86 \\
en-hi & 0.62 & en-zh & 4.02 \\
\bottomrule
\end{tabular}
\caption{Parallel data used for pre-training.}
\label{table:parallel:data}
\end{table}

\section{Results of Training From Scratch}
\label{sec:from:scratch}

\begin{table}[t]
\centering
\small
\begin{tabular}{lcccc}
\toprule
\textbf{Model} & \textbf{XNLI} & \textbf{MLQA} \\
Metrics & Acc. & F1 / EM \\ \midrule
MMLM$_\textsc{Scratch}$ & 69.40 & 55.02 / 37.90 \\
\our{}$_\textsc{Scratch}$ & \textbf{70.71} & \textbf{59.71} / \textbf{41.46} \\
~~$-$\xlco{} & 70.64 & 57.70 / 40.21 \\
~~$-$TLM & 69.76 & 58.22 / 40.78 \\ 
~~$-$MMLM & 63.06 & 52.81 / 35.01 \\ 
\bottomrule
\end{tabular}
\caption{Ablation results of the models pre-trained from scratch. Results are averaged over five runs.}
\label{table:fs-results}
\end{table}

We conduct experiments under the setting of training from scratch.
The Transformer size and hyperparameters follow BERT-base~\cite{bert}.
The parameters are randomly initialized from $U[-0.02, 0.02]$.
We optimize the models with Adam using a batch size of $256$ for a total of $1$M steps. The learning rate is scheduled with a linear decay with 10K warmup steps, where the peak learning rate is set as $0.0001$.
For cross-lingual contrast, we set the queue length as $16,384$.
We use a warmup of $200$K steps for the key encoder and then enable cross-lingual contrast.
We use an inverse square root scheduler to set the momentum coefficient, i.e., $m = \min(1-t^{-0.51}, 0.9995)$, where $t$ is training step.

Table~\ref{table:fs-results} shows the results of $\our{}_\textsc{Scratch}$ and various ablations.
$\our{}_\textsc{Scratch}$ significantly outperforms MMLM$_\textsc{Scratch}$ on both XNLI and MLQA.
We also evaluate the pre-training objectives of \our{}, where we ablate \xlco{}, TLM, and MMLM, respectively.
The findings agree with the results in Table~\ref{table:ablations}.

\section{Hyperparameters for Pre-Training}

As shown in Table~\ref{table:pt-hparam}, we present the hyperparameters for pre-training \our{}. We use the same vocabulary with XLM-R \cite{xlmr}.

\begin{table}[ht]
\centering
\scriptsize
\renewcommand\tabcolsep{2.8pt}
\begin{tabular}{lrrr}
\toprule
Hyperparameters & \textsc{From Scratch} & \textsc{Base} & \textsc{Large} \\ \midrule
Layers & 12 & 12 & 24 \\
Hidden size & 768 & 768 & 1,024 \\
FFN inner hidden size & 3,072 & 3,072 & 4,096 \\
Attention heads & 12 & 12 & 16 \\
Training steps & 1M & 150K & 200K \\
Batch size & 256 & 2,048 & 2,048 \\
Adam $\epsilon$ & 1e-6 & 1e-6 & 1e-6 \\
Adam $\beta$ & (0.9, 0.999) & (0.9, 0.98) & (0.9, 0.98) \\
Learning rate & 1e-4 & 2e-4 & 1e-4 \\
Learning rate schedule & Linear & Linear & Linear \\
Warmup steps & 10,000 & 10,000 & 10,000 \\
Gradient clipping & 1.0 & 1.0 & 1.0 \\
Weight decay & 0.01 & 0.01 & 0.01 \\
Momentum coefficient & 0.9995* & 0.9999 & 0.999 \\
Queue length & 16,384 & 131,072 & 131,072 \\
Universal layer & 8 & 8 & 12 \\
\bottomrule
\end{tabular}
\caption{Hyperparameters used for \our{} pre-training. *: the momentum coefficient uses an inverse square root scheduler $m=\min(1-t^{-0.51}, 0.9995)$.}
\label{table:pt-hparam}
\end{table}

\section{Hyperparameters for Fine-Tuning}

In Table~\ref{table:hparam1} and Table~\ref{table:hparam2}, we present the hyperparameters for fine-tuning on XNLI and MLQA. For each task, the hyperparameters are searched on the joint validation set of all languages (\#M=1). For XNLI, we evaluate the model every 5,000 steps, and select the model with the best accuracy score on the validation set. For MLQA, we directly use the final learned model. The final scores are averaged over five random seeds.

\begin{table}[t]
\centering
\scriptsize
\begin{tabular}{lrr}
\toprule
& XNLI & MLQA \\ \midrule
Batch size & 32 & \{16, 32\} \\
Learning rate & \{5e-6, 7e-6, 1e-5\} & \{2e-5, 3e-5, 5e-5\} \\
LR schedule & Linear & Linear \\
Warmup & 12,500 steps & 10\% \\
Weight decay & 0 & 0 \\
Epochs & 10 & \{2, 3, 4\} \\
\bottomrule
\end{tabular}
\caption{Hyperparameters used for fine-tuning \textsc{BASE}-size models on XNLI and MLQA.}
\label{table:hparam1}
\end{table}

\begin{table}[t]
\centering
\scriptsize
\begin{tabular}{lrr}
\toprule
& XNLI & MLQA \\ \midrule
Batch size & 32 & 32 \\
Learning rate & \{4e-6, 5e-6, 6e-6\} & \{2e-5, 3e-5, 5e-5\} \\
LR schedule & Linear & Linear \\
Warmup & 5,000 steps & 10\% \\
Weight decay & \{0, 0.01\} & 0 \\
Epochs & 10 & \{2, 3, 4\} \\
\bottomrule
\end{tabular}
\caption{Hyperparameters used for fine-tuning \textsc{LARGE}-size models on XNLI and MLQA.}
\label{table:hparam2}
\end{table}

\end{document}

%% file: settings.tex
\usepackage{multirow}
\usepackage{amsmath}
\usepackage{capt-of}
\usepackage{tabularx}
\usepackage[caption=false]{subfig}
\usepackage{epsfig}
\usepackage{amssymb}
\usepackage{amsfonts}
\usepackage{booktabs}
\usepackage{scalerel}
\usepackage[inline]{enumitem}
\usepackage{listings}
\usepackage{varwidth}
\usepackage[export]{adjustbox}
\usepackage{tikz}
\usetikzlibrary{tikzmark}
\usepackage{cleveref}

\usepackage{stmaryrd}
\usepackage{bbm}

\newcommand{\sptk}[1]{\texttt{[#1]}}
\newcommand{\eqform}[1]{Equation~(\ref{#1})}

\usepackage{algorithm}
\usepackage[noend]{algpseudocode}

\definecolor{deepblue}{rgb}{0,0,0.5}
\definecolor{officeblue}{RGB}{0,102,204}
\definecolor{deepred}{rgb}{0.6,0,0}
\definecolor{deepgreen}{rgb}{0,0.5,0}
\definecolor{mybrickred}{RGB}{182,50,28}

\definecolor{fillcolor}{RGB}{216,217,252}

\algnewcommand\algorithmicrequireb{{\hspace{0.85cm}}}
\algnewcommand\INPTDESCB{\item[\algorithmicrequireb]}

\algnewcommand\algorithmicfuncdesc{\textbf{Function:}}
\algnewcommand\FUNCDESC{\item[\algorithmicfuncdesc]}
\algnewcommand\algorithmicfuncdescb{{\hspace{1.48cm}}}
\algnewcommand\FUNCDESCB{\item[\algorithmicfuncdescb]}
\algnewcommand{\algorithmicgoto}{\textbf{goto}}
\algnewcommand{\Goto}[1]{\algorithmicgoto~\ref{#1}}

%% file: math_commands.tex
\usepackage{amsmath,amsfonts,bm}

\def\eqref#1{equation~\ref{#1}}
\def\1{\bm{1}}

\def\vtheta{{\bm{\theta}}}

\DeclareMathAlphabet{\mathsfit}{\encodingdefault}{\sfdefault}{m}{sl}
\SetMathAlphabet{\mathsfit}{bold}{\encodingdefault}{\sfdefault}{bx}{n}

\newcommand{\Ls}{\mathcal{L}}